\documentclass[conference]{IEEEtran}

\IEEEoverridecommandlockouts
\usepackage{cite}
\usepackage{amsmath,amssymb,amsfonts}
\usepackage{algorithmic}
\usepackage{graphicx}
\usepackage{textcomp}
\usepackage{xcolor}

\usepackage{latexsym}
\usepackage{booktabs}
\usepackage{multirow}
\usepackage{microtype}
\usepackage{inconsolata}
\usepackage{colortbl}
\usepackage{subfig}
\usepackage{paralist}
\usepackage{comment}
\usepackage{url}
\usepackage[export]{adjustbox}
\usepackage{wrapfig}
\usepackage{colortbl}
\definecolor{lightgray}{gray}{0.9}

\def\BibTeX{{\rm B\kern-.05em{\sc i\kern-.025em b}\kern-.08em
    T\kern-.1667em\lower.7ex\hbox{E}\kern-.125emX}}
\begin{document}

\title{ Can Large Language Models Generate Effective Datasets for Emotion Recognition in Conversations?  \\

}
\author{Burak Can Kaplan, Hugo Cesar De Castro Carneiro, Stefan Wermter}

%\thanks{B. C. Kaplan, H. C. C. Carneiro and S. Wermter are with the Knowledge Technology Group, Department of Informatics, University of Hamburg.}

\author{
\IEEEauthorblockN{Burak Can Kaplan\IEEEauthorrefmark{1}, Hugo Cesar De Castro Carneiro\IEEEauthorrefmark{1}, and Stefan Wermter\IEEEauthorrefmark{1}}
\IEEEauthorblockA{\IEEEauthorrefmark{1}Department of Informatics, University of Hamburg, Hamburg 22527, Germany\\
}
}

\maketitle

\begin{abstract}

Emotion recognition in conversations (ERC) focuses on identifying emotion shifts within interactions, representing a significant step toward advancing machine intelligence. However, ERC data remains scarce, and existing datasets face numerous challenges due to their highly biased sources and the inherent subjectivity of soft labels. Even though Large Language Models (LLMs) have demonstrated their quality in many affective tasks, they are typically expensive to train, and their application to ERC tasks—particularly in data generation—remains limited. To address these challenges, we employ a small, resource-efficient, and general-purpose LLM to synthesize ERC datasets with diverse properties, supplementing the three most widely used ERC benchmarks. We generate six novel datasets, with two tailored to enhance each benchmark. We evaluate the utility of these datasets to (1) supplement existing datasets for ERC classification, and (2) analyze the effects of label imbalance in ERC. Our experimental results indicate that ERC classifier models trained on the generated datasets exhibit strong robustness and consistently achieve statistically significant performance improvements on existing ERC benchmarks.

\end{abstract}

\begin{IEEEkeywords}
large language models, machine learning, data generation, affective computing
\end{IEEEkeywords}

\section{Introduction}
\label{sec:intro}

Emotion recognition in conversations (ERC) is a relatively new field of study that focuses on identifying and understanding human emotions expressed during interactions~\cite{poria2019emotion}. Its primary goal is to detect emotion shifts within dialogues, a capability that has become increasingly important with the rise of social robotics and applications requiring emotionally intelligent systems~\cite{MAHDI2022104193}. Large Language Models (LLMs) have demonstrated substantial improvements in various natural language processing (NLP) tasks, including affective computing~\cite{amin2023can}, and hold potential as effective tools for generating ERC data. Despite their success, most evaluations of LLMs in affective tasks have been conducted with API-based models, which are expensive, or top-performing local models requiring significant computational resources~\cite{amin2023affectiv,lei2023instructerc}. Exploring the capabilities of small and general-purpose LLMs in ERC tasks thus emerges as a promising and cost-efficient alternative.

A critical challenge in ERC lies in the scarcity of high-quality, diverse datasets. Most existing datasets are derived from biased sources such as scripted TV shows or social media, which often feature imbalanced label distributions~\cite{poria-etal-2019-meld,zahiri2018emotion}. Moreover, crafting such datasets is costly and time-consuming due to participant recruitment, ethical concerns, unbiased dialogue construction, and the difficulty of accurate and consistent labeling. Annotating emotional data often involves subjective interpretations, with annotators frequently providing inconsistent labels for the same utterance. Typically, a majority vote is used to select a label when there is disagreement, but this process is limited by the small number of annotators (usually 3–5), leading to reliability issues~\cite{zahiri2018emotion,busso2008iemocap}. Additionally, as highlighted in Tab.~\ref{tab:limits}, existing ERC datasets vary significantly in their emotion label sets, speaker numbers, and languages, making it difficult to combine them effectively for transfer learning. Furthermore, the inconsistencies in emotion categories across datasets limit their interoperability, leading some studies to rely on mappings based on psychological studies to align these different label sets~\cite{lei2023instructerc}. However, such mappings are often rough approximations, raising concerns about their accuracy and applicability. Rooted in the concept that only the entity expressing a particular affective state can fully recognize it~\cite{picard2000affective}, we hypothesize that by having the LLM generate both utterances and their corresponding emotion labels simultaneously, we can address these issues and significantly improve dataset consistency. Tab.~\ref{tab:limits} highlights the characteristics of popular ERC datasets and underscores their limitations.

\begin{table}[t!]
\caption{
ERC datasets
}\label{tab:limits}
\renewcommand{\arraystretch}{1}
\setlength\tabcolsep{5pt}
    \centering
    \begin{tabular}{lcccc}
    \toprule
        \textbf{Dataset}     & \textbf{Source} & \textbf{Speaker} & \textbf{Emotions} & \textbf{Language} \\ \midrule 
\textbf{MELD}~\cite{poria-etal-2019-meld}         & TV              & Multiple               & 7                 & English           \\
\textbf{IEMOCAP}~\cite{busso2008iemocap}                & Scripted        & 2                & 6                 & English           \\
\textbf{EmoryNLP}~\cite{zahiri2018emotion}                   & TV              & 6                & 7                 & English           \\
\textbf{DailyDialog}~\cite{li2017dailydialog}                  & Scripted        & 2                & 7                 & English           \\  
\textbf{CPED}~\cite{chen2022cped}                 & TV              & Multiple                & 13                & Chinese           \\
\textbf{EC}~\cite{chatterjee2019semeval}                  & Twitter         & 3                & 4                 & English           \\
\textbf{KDEmor}\cite{pant2022korean}                  & TV              & Multiple                & 3                 & Korean            \\
        \bottomrule
    \end{tabular}
    
\end{table}

\begin{figure}[ht!]
    \centering
    \subfloat{\includegraphics[width=0.432\textwidth]{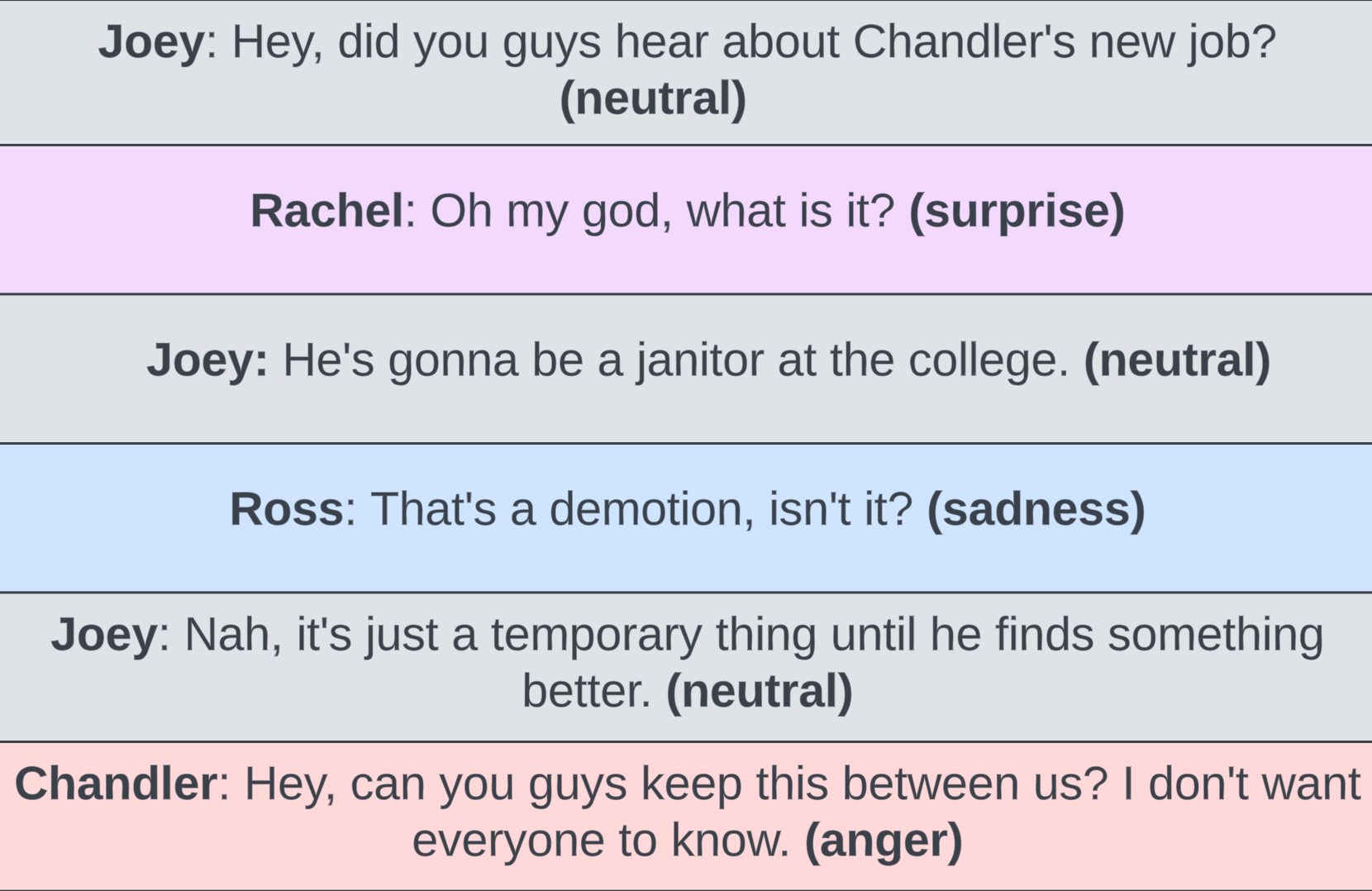}}\hfill
        \caption{Example of a generated dialogue, featuring multiple speakers, utterances, and corresponding emotion labels.}
            \label{fig:sample}
\end{figure}

To tackle these limitations, we propose leveraging a small, general-purpose LLM to synthesize new ERC datasets. By generating both dialogue lines and their corresponding emotion labels in a single step, we aim to improve the consistency and reliability of ERC data while avoiding the costs and complexities associated with traditional data collection and annotation methods. To ensure comparability with existing benchmarks, we generate six new datasets, with two corresponding datasets for each of the three widely used ERC benchmarks. This alignment allows us to systematically evaluate the potential of LLM-generated datasets to supplement existing resources, mitigate label imbalance, and enhance ERC model performance. Fig.~\ref{fig:sample} shows an example of a dialogue generated by our LLM.

Our contributions are as follows: Firstly, we demonstrate the capability of a small LLM to generate consistent, multi-party ERC datasets suitable for training ERC models. Secondly, we propose a methodology to evaluate the quality and validity of synthesized ERC data, focusing on their utility in improving model robustness and performance. Thirdly, we assess the effects of different label distributions in ERC datasets and analyze how these imbalances affect existing benchmarks. The prompts and parameters used for data generation are shared in the paper, ensuring reproducibility with local LLMs. Additionally, the code repository will be made available to facilitate further advancements in LLM-based ERC dataset creation.

The remainder of the paper is organized as follows: Section~\ref{sec:related} reviews existing ERC datasets and related research on LLMs for affective computing, dataset enhancement, and synthetic data generation. Section~\ref{sec:generation} details the LLM setup, the dataset synthesis process, prompt engineering, and all parameters used, providing all necessary information for transparency. Section~\ref{sec:results} presents experimental evaluations of the generated datasets, and Section~\ref{sec:conclusion} concludes with insights and directions for future research.

\section{Related Work}

\label{sec:related}

\subsection{Existing Datasets}
\label{sec:datasets}

In this study, we focus on the three most popular ERC datasets in Papers With Code\footnote{\url{https://paperswithcode.com/task/emotion-recognition-in-conversation}}: MELD~\cite{poria-etal-2019-meld}, EmoryNLP~\cite{zahiri2018emotion}, and IEMOCAP~\cite{busso2008iemocap}. These datasets are chosen to assess the performance of ERC models on our generated datasets, as well as on these existing datasets.

\textbf{MELD} is a dataset constructed by extracting lines from the ``Friends'' TV series. It encompasses 7 emotions: Neutral, Disgust, Anger, Sadness, Fear, Joy, and Surprise. There is a high imbalance among its labels, with Neutral being the most common. On the other hand, Disgust and Fear are notably rare. In some studies, authors perform classification task without either one or both of these labels. According to Papers With Code, weighted F1 classification results for MELD usually lie within the range from 60\% to 70\%.

The \textbf{EmoryNLP} dataset is also constructed from  the lines of ``Friends'' TV series, but it employs different emotion labels than MELD, which are: Sad, Mad, Scared, Powerful, Peaceful, Joyful, and Neutral. For the annotation process, 4 annotators participated, and only 6.17\% of the annotations correspond to labels in which there was an unanimous agreement among the annotators. In 9.39\% of the annotations, each annotator labeled the same data with different label. Majority voting was used to select the annotation in most cases. The weighted F1 results for EmoryNLP, as reported on Papers With Code, are around 35\%, which is significantly lower compared to other datasets, characterizing EmoryNLP as a challenging dataset.

\textbf{IEMOCAP} is a scripted dyadic dataset which also provides multimodal information. It encompasses 10 labels: Neutral, Happiness, Sadness, Anger, Excited, Frustration, Fear, Surprise, Disgust, and Other (Uninformative). Disgust does not show up in any record in the conversations of the validation split. Classification tasks on IEMOCAP involve the utilization of specific subsets of its available classes. Those that do not include Disgust and Other are denoted 8-way. Fear and Surprise are very rare throughout the dataset, so papers often do not include them. This classification task is denoted 6-way. Finally, although with more observations, Excited and Frustration appear considerably less than Neutral, Happiness, Sadness, and Anger. A classification using solely these 4 labels is denoted 4-way. Results aggregated on Papers With Code reveal that weighted F1 scores for IEMOCAP are slightly higher than those for MELD regardless of the number of classes used. This paper adopts the 6-way classification, as it is the most commonly employed approach in the literature.

\subsection{LLM Usage}

MELD, IEMOCAP, and EmoryNLP are multimodal datasets, but they are often exclusively benchmarked in the textual modality~\cite{lei2023instructerc,song-etal-2022-supervised} due to the inherent noise in their audio and visual components, which poses challenges to ERC model training. Even though some approaches aim to enhance audio-visual data quality through preprocessing~\cite{pohjalainen2016spectral,carneiro2023whose} or modifying the modality fusing approach~\cite{hu2022mm,zhang-li-2023-cross,10650481}, opportunities for improvement in this field still persist. Language models are frequently leveraged to increase the quality of existing datasets as well. For instance, pretrained language models can provide additional information at the utterance and conversation levels, thus increasing the context of an existing ERC dataset~\cite{kim-etal-2020-contextual}. Additionally, LLMs are also used in annotation process of affective speech data to enhance its quality~\cite{latif2023large,amin2023affectiv}. 

\begin{figure*}[t!]
    \centering
    \subfloat[][MELD]{\includegraphics[height=0.275\textheight]{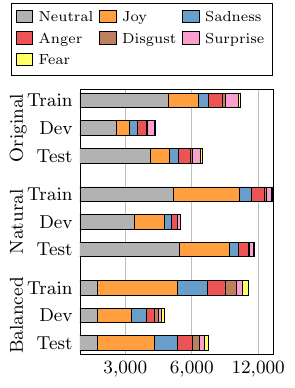}\label{fig:label_distribution_meld}}\hfill
    \subfloat[][EmoryNLP]{\includegraphics[height=0.275\textheight]{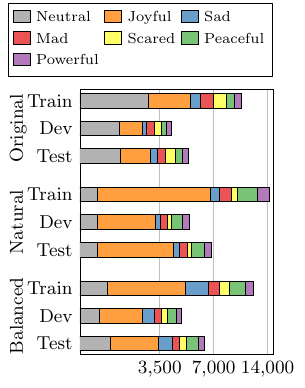}\label{fig:label_distribution_emory}}\hfill
    \subfloat[][IEMOCAP]{\includegraphics[height=0.275\textheight]{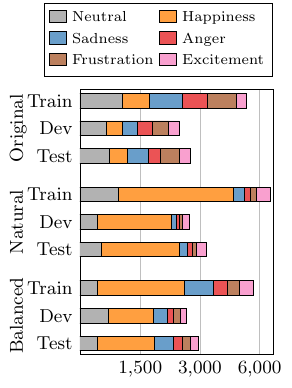}\label{fig:label_distribution_iemo}}
    \caption{Label distribution of reference and generated datasets. The horizontal axis represents the number of utterances within each dataset, and is shown in a logarithmic scale to enhance the visualization of rare labels.}
    \label{fig:label_distribution}
\end{figure*}

Empathetic intelligence also requires measuring the capabilities of an LLM in various affective tasks. Existing literature indicates that LLMs exhibit considerable empathetic intelligence. Zhu et al.~\cite{zhu-etal-2021-topic} demonstrate that transformer-based architectures are capable of distinguishing emotions in dialogues, and Deng et al.~\cite{10650044} use transformer architectures to generate dialogue responses. Amin et al~\cite{amin2023affectiv} assess ChatGPT in affective computing, revealing that even a generalized model can achieve decent results.  ChatGPT's knowledge has been also measured in solving affective computing problems, namely sentiment analysis, personality assessment, and suicide tendency detection~\cite{amin2023can}. LLMs are also evaluated in tasks like affective support, multi-party conversations and  ERC~\cite{zhao2023chatgpt,tan-etal-2023-chatgpt}. Tu et al.~\cite{tu2023empirical} extract detailed additional knowledge from ERC data using ChatGPT and measures its impact on ERC models. Feng et al.~\cite{feng2023affect} show that LLMs can serve as effective classifiers for affect recognition in conversation tasks. Additionally, LLM-based models have achieved high classification scores on widely used ERC datasets~\cite{lei2023instructerc}, further supporting the suitability of LLMs for ERC data generation.

There are approaches to enhance the LLMs' data generation capabilities. Eldan et al.~\cite{eldan2023tinystories} employ GPT-3.5 and GPT-4 to generate child-level language to train small language models. Josifoski et al.~\cite{josifoski-etal-2023-exploiting} present a synthetic data generation pipeline that involves prompting LLMs to generate text from coherent triplets extracted from a knowledge graph. Conversely, Chung et al.~\cite{chung2023increasing} attempt to increase the diversity of LLM data, acknowledging potential trade-offs with lower output accuracy. Veselovsky et al.~\cite{veselovsky2023generating} use LLM synthetic data to train classifiers, evaluating them on real data with various strategies. In summary, the existing literature underscores LLMs' effectiveness in diverse affective tasks. However, despite the significant limitations and noise present in existing ERC datasets, research on leveraging LLMs specifically for ERC data generation remains scarce, leaving an open opportunity for further exploration in this direction.

\section{Dataset Generation}
\label{sec:generation}

In the LLM selection phase, we ran small dialogue generation experiments, and observed that 7 billion-sized models proved incapable of generating creative and diverse dialogues while keeping sufficient output consistency for use as a dataset, often exhibiting repetitions of words or sentences. Regarding the larger models, we decided not to use ChatGPT despite yielding the most favorable results, due to our emphasis on ensuring the reproducibility of our approach. Although 33 billion-sized models yielded decent results, dialogues from the 13 billion-sized model appeared natural and required roughly 25 GB VRAM. Thus, we opted for using a small model with a reasonable GPU to offer an affordable and computation-efficient solution for ERC dataset generation, with Vicuna 1.5\footnote{\url{https://huggingface.co/lmsys/vicuna-13b-v1.5}} being the 13 billion-sized model to provide the best and most consistent results, and one of the most popular open sourced LLMs.

\subsection{Natural and Balanced Data}
\label{sec:natbaldata}
To assess our local LLMs' capabilities in generating multi-party affective conversations across various aspects, we generated two types of dataset with the same set of labels and dialogue structure of the three mentioned datasets in Sec.~\ref{sec:datasets}, providing a total of 6 new datasets. These dataset types are named ``Natural'' and ``Balanced''. Each dataset was intentionally over-generated to safeguard against data limitations and to facilitate comparison with their corresponding original datasets. Fig.~\ref{fig:label_distribution} displays total utterances, label amounts and label distributions of each dataset.

Natural datasets comprise dialogues created freely by an LLM without any predetermined bias in its generation process. These datasets show the broad potential of LLMs in creating dialogues, with the distribution of the emotion labels within them being closer to reality. For example, emotions like happiness, sadness, or even the absence of emotions being far more common than emotions like fear and surprise. The generation of natural data is particularly useful for the development of affective interactive systems, e.g., more realistic and natural social robots to be used in real-life applications. With natural datasets, we aim to assess how naturally LLMs generate dialogues without emotional pre-conditioning. 

Balanced datasets, conversely, comprise dialogues with more evenly distributed emotions and they are designed to address class imbalance in existing ERC datasets while maintaining natural dialogue flow. To preserve the naturalness of the dialogues, we instruct the LLM to generate conversations that include at least one utterance with a specific given emotion. However, the content and placement of these within the dialogue, as well as the emotions of the remaining utterances, are left for the LLM to decide. This procedure does not yield a dataset where all emotions are uniformly distributed; rather, it ensures that a particular emotion appears in a significant number of dialogues. Fig~\ref{fig:label_distribution} shows that the balanced datasets have the most balanced label distributions among all datasets. Due to the intentional bias introduced in its generation process, balanced datasets are not well suited for generative and affective usage, since rare emotions may appear more often than they do in real-life conversations. Nevertheless, these datasets are better fitting in the development of more accurate classifiers.

\begin{figure}[t]
    \centering
    \subfloat{\includegraphics[width=0.485\textwidth]{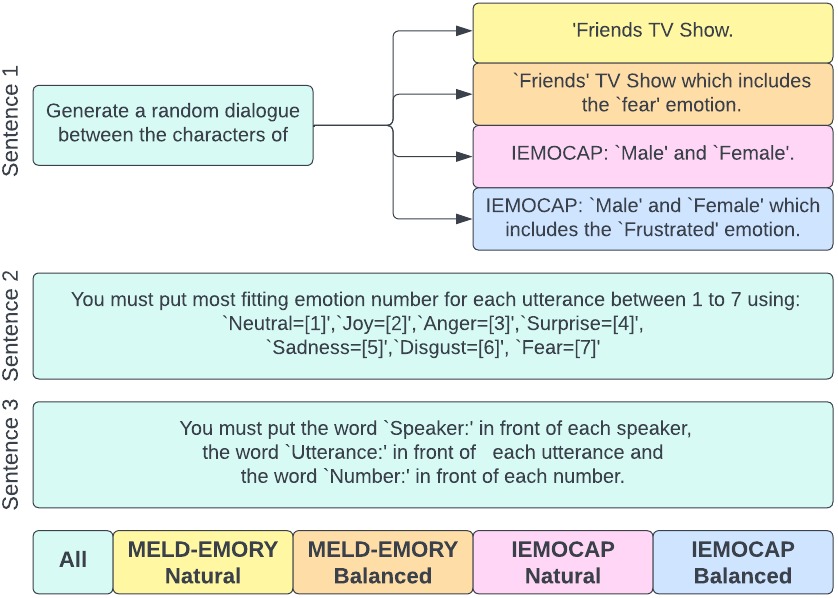}}\hfill
        \caption{
Prompt examples for each generated dataset. Each prompt consists of three sentences (1:Task Definition, 2:Labelling with Logic Reasoning, 3:Structuring). The emotion labels specified in the diagram are used for illustration and are modified with matching labels for each dataset being generated.}\label{fig:prompttable}
         
\end{figure}

\subsection{Prompt Engineering}
\label{sec:prompteng}
Prompting is a critical aspect of this research, directly biasing the LLM towards the goal. In the generation of natural data, four tasks were needed to achieve a proper output which are providing speaker names, utterances, consistent emotion labels and structure. Due to the varying nature of the speakers in the target datasets, distinct prompts tailored to each dataset were employed. Specifically, IEMOCAP involves dyadic interactions with unnamed male and female participants, while EmoryNLP and MELD are derived from conversations in the ``Friends'' TV show.

The most challenging aspect of the prompting process was obtaining accurate emotion labels from the LLM, given that these models tend to hallucinate and often forget prompt details when faced with complex tasks, leading to inconsistencies. To restrict the LLM to generate emotion labels from a specific label group, we assign each emotion label with a number as symbols and ask the LLM to provide one of them for each utterance. In the literature, the same logic is used in LLM Logic Reasoning~\cite{10.24963,creswell2022faithful,qiao-etal-2023-reasoning}, and it is a proven technique that improves the LLM performance. In our case, this method ensures the generation of consistent labels across datasets.

The last task involved employing a prompt to generate structured outputs to facilitate the retrieval of the necessary data. To ensure the parseability of that structure, we instructed the LLM to prepend speaker, utterance, and number (representing emotion) to the corresponding data. Fig.~\ref{fig:prompttable} provides the prompts used to generate the dialogues with this structure. At the generation, all sentences are concatenated and submitted to the LLM as a single prompt. Fig.~\ref{fig:sample} displays an example of ERC dialogue generated by the LLM.

For the generation of balanced data, the prompt is kept the same, except for the last part of the first sentence, where we instruct the LLM to ensure the dialogue contains at least one utterance expressing a specific emotion. Fig.~\ref{fig:prompttable} offers an example of prompt used to ensure the existence of at least one utterance expressing fear. To preserve the naturalness of the dialogue, we restrain from specifying the number associated with that label. This prompt iterates through all labels within the target dataset, ensuring an equal number of dialogues containing utterances expressing some particular emotion.

We tested these prompts with the LLM several times with the parameters provided in Tab.~\ref{tab:parametertable}. It was observed that the LLM consistently produced the same outputs, ensuring reproducibility. Specific words drive each LLM to produce different outputs based on their weights, and the prompts here offered were tailored for Vicuna 1.5(13b).

\begin{figure*}[t!]
    \centering
    \includegraphics[height=0.21\textheight]{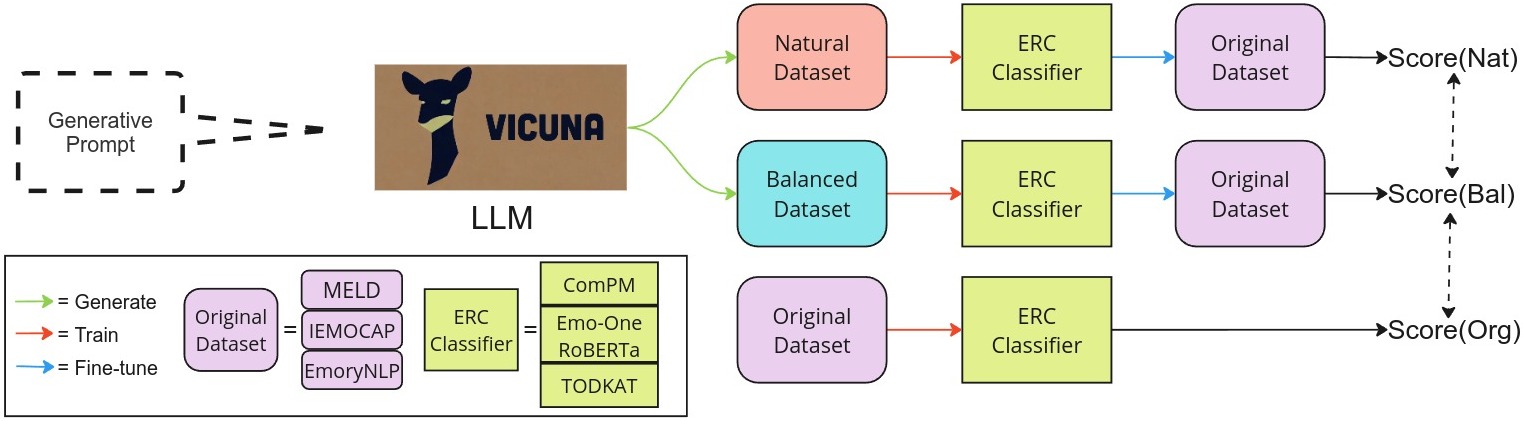}
    \caption{Diagram of our pipeline.}
    \label{fig:pipeline}
  %  \vspace{-6pt}
\end{figure*}

\subsection{LLM Parameters}
\label{sec:params}
For the utilization of the local LLM, we employed TextgenWebUI Chat API\footnote{\url{https://github.com/oobabooga/text-generation-webui}} with Langchain\footnote{\url{https://www.langchain.com/}}, which provides access to all LLM parameters. For reproducibility, the seed is randomized for diversity but fixed to maintain consistent random numbers for each LLM run. Tab.~\ref{tab:parametertable} shows all parameters used in our data generation process.

\begin{table}[]
\caption{LLM Parameters used. RP: Repetition Penalty}
\centering
\begin{tabular}{lllll}
\toprule
\small\textbf{Temperature} & \small\textbf{Top\_p} & \small\textbf{Top\_k} & \small\textbf{RP} & \small\textbf{Typ. p} \\
\multicolumn{1}{c}{\small0.7} & \multicolumn{1}{c}{\small1} & \multicolumn{1}{c}{\small10000} & \multicolumn{1}{c}{\small1}       & \multicolumn{1}{c}{\small0.995} \\ \bottomrule
\end{tabular}

\label{tab:parametertable}
\end{table}

Temperature is not crucial in our case due to the fixed seed, so we opted to keep it at its default value. We kept top p, top k, and typical p very high, to enhance context diversity and provide more options in the dialogues. Furthermore, we minimized repetition penalty to avoid limiting the LLM based on the tokens it produced.

\section{Experiments and Analyses}
\label{sec:results}

As outlined in Section~\ref{sec:natbaldata}, two datasets—natural and balanced—were generated for each of the datasets discussed in Section~\ref{sec:prompteng}. These datasets share the same speakers, structure, and emotion label set as their corresponding datasets for the purpose of comparison. To assess the utility of the generated datasets, we employed three popular ERC classifier architectures: CoMPM~\cite{lee-lee-2022-compm}, EmoOne-RoBERTa~\cite{lee2022emotion}, and TODKAT~\cite{zhu-etal-2021-topic}. These models were chosen based on their available implementations and competitive performance scores on all three datasets: MELD, EmoryNLP, and IEMOCAP. These architectures were preferred over some state-of-the-art models due to factors such as the lack of implementations for certain datasets or high VRAM requirements such as 4x80G nVidia A100~\cite{lei2023instructerc}. This decision aligns with our aim of delivering a reproducible and affordable solution. Furthermore, the primary focus of this research is not achieving the highest scores but demonstrating whether the datasets generated through our approach boost the ERC classification process. To further validate our findings, we conducted statistical tests on the experiment results from the three classifier architectures, ensuring the significance of our conclusions. The entire pipeline used for evaluation is illustrated in Fig.~\ref{fig:pipeline}.

\subsection{Assessing Synthetic Dataset Properties}
\label{sec:Assessment}

To evaluate the utility of the generated datasets for ERC and the methodology proposed in this paper, we need to assess whether these datasets possess desirable properties relevant to the task. These properties include:
\begin{inparaenum}[1.]
    \item having all their splits sampled from the same distribution;
    \item being able to train models to be robust to unseen data; and
    \item having the potential to pretrain models for better performance when fine-tuned with benchmark datasets.
\end{inparaenum}
The first property is guaranteed by our methodology since dialogues are independently generated by the LLM using the same prompt and parameters. As dialogues are independently generated and the dataset is only split afterward, all generated datasets have splits sampled from the same distribution. The second property requires training ERC architectures on the generated datasets and subjecting them to unseen data. The third property involves training the same architectures on the generated datasets and further fine-tuning them on the corresponding reference dataset to evaluate whether it results in significant improvements in the recognition capability. 

To evaluate both robustness and fine-tuning potential, we split each generated dataset into training, validation, and test sets, ensuring that a portion of the data remained unseen for evaluation at later stages. Additionally, models were tested on unseen data by evaluating them on the original test splits of MELD, IEMOCAP, and EmoryNLP. In the literature, this evaluation strategy—train on synthetic, test on real (TSTR)—is commonly used to assess the effectiveness of synthetic data in downstream tasks and represents the most suitable evaluation scheme for our study~\cite{hyland2017real,yuan2024multifacetedevaluationframeworkassessing}.

After splitting the datasets, we trained the CoMPM, EmoOne-RoBERTa, and TODKAT architectures separately using the training splits of the generated datasets. Once training was completed, we fine-tuned all models on the training splits of MELD, IEMOCAP, and EmoryNLP for domain adaptation. Finally, we evaluated the fine-tuned models on the test splits of MELD, IEMOCAP, and EmoryNLP to assess:
\begin{inparaenum}[1.]
\item whether they function as robust ERC classifiers, and
\item whether models trained on synthetic data exhibit improved performance compared to their original versions. The results of this experiment can be seen in Tab.~\ref{tab:results}
\end{inparaenum}

An initial review of Table~\ref{tab:results} shows that models trained on LLM-generated datasets exhibit strong robustness and generalization capabilities, producing ERC models that perform comparably to or better than those trained on original datasets. Most importantly, across all classifier architectures, models trained on generated datasets (highlighted in gray backgrounds) consistently outperform their original counterparts across all three benchmarks. This demonstrates that even small LLM-generated datasets can enhance ERC classifier performance through transfer learning, thereby achieving the primary objective of this study.

\subsection{Assessment of the Effects of Different Label Distributions}

In this subsection, we compare model performances across different synthetic label distributions, as presented in Table~\ref{tab:results}. Surprisingly, our experiments reveal that not all benchmarks behave similarly when evaluating models trained on datasets with different label distributions. Balanced datasets yielded the highest scores on the MELD dataset across all classifiers, highlight that this dataset needs much more balanced labels to be more effective for ERC. In contrast, models trained on natural datasets performed best on IEMOCAP, suggesting that class imbalance is less critical for this benchmark and that training should prioritize data that reflects real-world distributions. For EmoryNLP, classification scores were consistently lower than in the other two benchmarks, indicating its greater complexity. Moreover, no clear advantage was observed between training on balanced versus natural datasets for this dataset. These findings suggest that the impact of label distributions on model performance is dataset-dependent, highlighting the importance of tailoring synthetic dataset generation strategies to the characteristics of individual benchmarks.

\begin{table}[t!]
    
    \caption{Performance results of three ERC classifiers across all three benchmarks (wf1). Org denotes scores from the original dataset, while generated dataset scores (Nat and Bal) are highlighted with a colored background for clearer distinction. The highest scores for each benchmark is marked in bold.}
    \centering
    \label{tab:results}
    \begin{tabular}{ccccc}
        \toprule
        \textbf{}  & \textbf{Test Set} & \textbf{ComPM} & \textbf{EmoOne} & \textbf{TODKAT} \\
        \midrule
        \multirow{3}{*}{\textbf{MELD}} 
            & Org  & 65.43 & 65.46 & 63.47 \\
            & Nat  & \cellcolor{lightgray}65.52 & \cellcolor{lightgray}66.50 & \cellcolor{lightgray}64.20 \\
            & Bal  & \cellcolor{lightgray}\textbf{66.16} & \cellcolor{lightgray}\textbf{67.27} & \cellcolor{lightgray}\textbf{64.27} \\
        \midrule
        \multirow{3}{*}{\textbf{EMORYNLP}}  
            & Org  & 37.25 & 35.93 & 35.38 \\
            & Nat  & \cellcolor{lightgray}\textbf{39.50} & \cellcolor{lightgray}38.79 & \cellcolor{lightgray}36.77 \\
            & Bal  & \cellcolor{lightgray}38.93 & \cellcolor{lightgray}\textbf{39.05} & \cellcolor{lightgray}\textbf{37.40} \\
        \midrule
        \multirow{3}{*}{\textbf{IEMOCAP}}  
            & Org  & 65.21 & 67.19 & 54.63 \\
            & Nat  & \cellcolor{lightgray}\textbf{68.06} & \cellcolor{lightgray}\textbf{69.28} & \cellcolor{lightgray}\textbf{55.96} \\
            & Bal  & \cellcolor{lightgray}67.87 & \cellcolor{lightgray}67.81 & \cellcolor{lightgray}53.39 \\
        \bottomrule
    \end{tabular}
\end{table}

\subsection{Further Validation on Findings}

To further validate our findings, we conducted an additional experiment, where we subjected the models used in Section~\ref{sec:Assessment} to unseen test data split from the generated datasets additionally. The models were then ranked across all test sets and architectures to enable a comparative analysis.

We performed a Friedman rank sum test~\cite{friedman1937use}, which is a non-parametric statistical test that has been established as a scientifically valid way to evaluate the significant improvement of a classifier in comparison to several others over various datasets~\cite{demvsar2006statistical}. Due to the relatively small number of classifiers (9) and datasets (9), we could not resort to chi-squared approximations and calculated the exact $p$-values using the formula introduced by Eisinga et al.~\cite{eisinga2017exact}. 

The Friedman test is non-parametric, as the dataset does not follow a normal distribution. Consequently, ranking-based analysis is applied instead of using raw W-F1 scores. Each trained model instance—whether trained solely on the original dataset or pretrained on synthetic data and subsequently fine-tuned on the original dataset—was assigned a rank from 1 (highest W-F1 score) to 9 (lowest W-F1 score). This ranking procedure was repeated for each test set, corresponding to each row in Table~\ref{tab:results_experiment_finetuning}.

After ranking, the rank sums for each model across all test sets were calculated. A lower rank sum indicates consistently strong performance, while a higher rank sum suggests weaker performance across the test sets. The rank sum of models pretrained on synthetic datasets (and fine-tuned on the original dataset) was then subtracted from the rank sum of models trained solely on the original dataset. If the rank sum of a model pretrained on synthetic data was lower than that of a model trained solely on the original dataset, this indicated that the pretrained model exhibited stronger overall performance. Conversely, if the rank sum was higher, the original model outperformed the pretrained one. Significant differences in rank sums suggested consistent performance disparities, which could indicate statistical significance.

Bonferroni-corrected $p$-values were calculated to provide a quantitative measure of statistical significance~\cite{dunn1961multiple}. Table~\ref{tab:results_experiment_finetuning} presents each model’s rank (in parentheses), rank sums, and absolute differences. The bottom row of the table provides the corresponding Bonferroni-corrected $p$-values.

The low $p$-values obtained through the Friedman rank sum test suggest that it is highly unlikely that the performance improvements of models pretrained on synthetic data are due to random chance. This effect is especially evident in CoMPM and EmoOne-RoBERTa, which demonstrated the strongest performance across benchmarks. However, no statistical significance was observed for TODKAT, likely due to inherent limitations of this model. These results further reinforce the potential of LLM-generated datasets in enhancing ERC classifier performance while highlighting variations in model-specific responses to synthetic pretraining.

\begin{table*}[ht!]

\caption{Extended version of results(W-F1), in which the test results of test splits of generated datasets and Friedman test calculations included. The W-F1 scores are ranked row-wise (1 for the highest, 9 for the lowest), with the highest in bold. The ``Rank'' part shows the sum of ranks and the absolute difference between rank sums compared to the original dataset-trained counterpart. The ``p'' row provides the $p$-values from the Friedman rank sum test.}
\label{tab:results_experiment_finetuning}
\centering
\begin{tabular}{p{1.5cm}p{0.7cm}  c@{\hskip 0.3cm}c@{\hskip 0.3cm}c@{\hskip 1.4cm}c@{\hskip 0.3cm}c@{\hskip 0.3cm}c@{\hskip 1.4cm}c@{\hskip 0.3cm}c@{\hskip 0.3cm}c}
 \toprule
   \multicolumn{1}{c}{\multirow{2}{*}{}} & \multicolumn{1}{c}{\multirow{2}{*}{\hspace{-0.2cm}\textbf{Test Set}}} & \multicolumn{3}{c}{\hspace{-1cm}\textbf{ComPM}} & \multicolumn{3}{c}{\hspace{-1.2cm}\textbf{EmoOne-RoBERTa}} & \multicolumn{3}{c}{\textbf{TODKAT}}  \\
 \cline{3-11}
  \multicolumn{2}{c}{} & \multicolumn{1}{c}{\textbf{Org}} & \multicolumn{1}{c}{\textbf{Nat}} & \multicolumn{1}{l}{\textbf{Bal}}  & \multicolumn{1}{c}{\textbf{Org}} & \multicolumn{1}{c}{\textbf{Nat}} & \multicolumn{1}{l}{\textbf{Bal}}  & \multicolumn{1}{c}{\textbf{Org}} & \multicolumn{1}{c}{\textbf{Nat}} & \multicolumn{1}{l}{\textbf{Bal}}\\
 \midrule
 \multirow{3}{*}{\textit{\textbf{MELD}}} & Org   & \small65.43$^{(6)}$ & \small65.52$^{(4)}$ & \small66.16$^{(3)}$  & \small65.46$^{(5)}$ & \small66.50$^{(2)}$ & \small\textbf{67.27}$^{(1)}$  & \small63.47$^{(9)}$ & \small64.20$^{(8)}$ & \small64.27$^{(7)}$  \\
  & Nat & \small48.07$^{(7)}$ & \small\textbf{50.96}$^{(1)}$ & \small50.29$^{(3)}$  & \small49.18$^{(6)}$ & \small50.95$^{(2)}$ & \small49.24$^{(5)}$  & \small46.52$^{(9)}$ & \small49.37$^{(4)}$ & \small47.86$^{(8)}$ \\
  & Bal  & \small58.66$^{(8)}$ & \small60.77$^{(6)}$ & \small65.99$^{(2)}$  & \small61.17$^{(5)}$ & \small61.20$^{(4)}$ & \small\textbf{66.10}$^{(1)}$  & \small57.34$^{(9)}$ & \small60.46$^{(7)}$ & \small62.30$^{(3)}$ \\
  \midrule
 \multirow{3}{*}{\textit{\textbf{EMORY-NLP}}} & Org   & \small37.25$^{(6)}$ & \small\textbf{39.50}$^{(1)}$ & \small38.93$^{(3)}$  & \small35.93$^{(8)}$ & \small38.79$^{(4)}$ & \small39.05$^{(2)}$  & \small35.38$^{(9)}$ & \small36.77$^{(7)}$ & \small37.40$^{(5)}$ \\
  & Nat & \small31.66$^{(6)}$ & \small35.89$^{(2)}$ & \small34.00$^{(5)}$  & \small28.85$^{(8)}$ & \small34.06$^{(4)}$ & \small34.73$^{(3)}$  & \small28.37$^{(9)}$ & \small\textbf{37.14}$^{(1)}$ & \small31.12$^{(7)}$ \\
  & Bal & \small47.67$^{(7)}$ & \small53.39$^{(4)}$ & \small60.86$^{(2)}$  & \small46.91$^{(8)}$ & \small50.51$^{(5)}$ & \small\textbf{60.92}$^{(1)}$  & \small38.15$^{(9)}$ & \small49.71$^{(6)}$ & \small56.95$^{(3)}$ \\
  \midrule
 \multirow{3}{*}{\textit{\textbf{IEMOCAP}}} & Org   & \small65.21$^{(6)}$ & \small68.06$^{(2)}$ & \small67.87$^{(3)}$  & \small67.19$^{(5)}$ & \small\textbf{69.28}$^{(1)}$ & \small67.81$^{(4)}$  & \small54.63$^{(8)}$ & \small55.96$^{(7)}$ & \small53.39$^{(9)}$ \\
  & Nat & \small16.76$^{(9)}$ & \small\textbf{37.58}$^{(1)}$ & \small27.08$^{(6)}$  & \small19.84$^{(8)}$ & \small35.85$^{(2)}$ & \small27.89$^{(5)}$  & \small26.02$^{(7)}$ & \small30.86$^{(3)}$ & \small30.27$^{(4)}$ \\
  & Bal & \small34.05$^{(8)}$ & \small53.89$^{(3)}$ & \small59.62$^{(2)}$  & \small33.20$^{(9)}$ & \small50.26$^{(4)}$ & \small\textbf{60.35}$^{(1)}$  & \small40.23$^{(7)}$ & \small41.94$^{(6)}$ & \small47.64$^{(5)}$ \\
  \midrule

 \multicolumn{2}{l}{\textit{\textbf{Rank}}} & \textbf{Org} & \textbf{Nat} & \textbf{Bal}  & \textbf{Org} & \textbf{Nat} & \textbf{Bal}  & \textbf{Org} & \textbf{Nat} & \textbf{Bal}\\
 \midrule
 & \textbf{Sum}   & \small63 & \small24  & \small29 & \small62 & \small28 & \small23 & \small76 & \small49 & \small51\\
 & \textbf{Diff.} & \small- & \small39 & \small34  & \small- & \small34 & \small39 & \small- & \small27 & \small25\\
 & \textbf{p} & \small- & \small\textbf{0.0034} & \small\textbf{0.0186} & \small- & \small\textbf{0.0186} & \small\textbf{0.0034} & \small- & \small0.1273 & \small0.2025\\
 \bottomrule
\end{tabular}

\end{table*}

\section{Conclusion}
\label{sec:conclusion}
This study introduces a reproducible, affordable, and computationally efficient approach for generating ERC datasets using a small, resource-efficient LLM with structured prompt engineering. By addressing key challenges such as soft labels, dataset incompatibilities, and class imbalance, our method enables scalable dataset creation without relying on expensive or black-box models. Our experimental results demonstrate that models trained on LLM-generated datasets exhibit enhanced recognition capabilities and improved performance on ERC benchmarks. We proposed a systematic approach to assessing the quality of these datasets, and confirmed their potential for fine-tuning ERC models. Statistical tests further solidify these findings, providing robust support for the impact of our approach. Additionally, we assessed the effects of having different label distributions in ERC, and our results highlight the need of having more balanced data for some particular benchmarks.

Looking ahead, fine-tuning existing LLMs (e.g., Llama 3) could enable the generation of even larger and more diverse affective datasets, further improving dataset adaptability. Additionally, given that scalability is one of the biggest advantages of LLM synthetic data, the research on its effects in ERC is also can be an interesting direction. Moreover, our methodology provides a foundation for generating customizable datasets not only for ERC but also for other NLP tasks. By releasing our parameters, code, and prompts, we aim to facilitate further research in synthetic data generation for affective computing and beyond.

%\section*{References}

%\small  % You can also try \footnotesize or \scriptsize
\bibliographystyle{ieeetr}
\bibliography{anthology,custom}
\end{document}